# Machine Learning for recognition of minerals from multispectral data

Pavel Jahoda[a,1] (Researcher), Igor Drozdovskiy[b] (Researcher), Francesco Sauro[b,c], Leonardo Turchi[b], Samuel Payler[b] and Loredana Bessone[b]

[a]*Czech Technical University in Prague, Zikova 1903/4, 166 36 Praha 6, Czechia*

[b]*European Space Agency, European Astronaut Centre, Linder Höhe, 51147, Cologne, Germany*

[c]*Department of Biological, Geological and Environmental Sciences, University of Bologna, Italy*

---

## ABSTRACT

Machine Learning (ML) has found several applications in spectroscopy, including being used to recognise minerals and estimate elemental composition. In this work, we present novel methods for automatic mineral identification based on combining data from different spectroscopic methods. We evaluate combining data from three spectroscopic methods: vibrational Raman scattering, reflective Visible-Near Infrared (VNIR), and Laser-Induced Breakdown Spectroscopy (LIBS). These methods were paired into Raman + VNIR, Raman + LIBS and VNIR + LIBS, and different methods of data fusion applied to each pair to classify minerals. The methods presented here are shown to outperform the use of a single data source by a significant margin. Additionally, we present a Deep Learning algorithm for mineral classification from Raman spectra that outperforms previous state-of-art methods. Our approach was tested on various open access experimental Raman (RRUFF) and VNIR (USGS, Relab, ECOSTRESS), as well as synthetic LIBS NIST spectral libraries. Our cross-validation tests show that multi-method spectroscopy paired with ML paves the way towards rapid and accurate characterization of rocks and minerals.

---

## 1. Introduction

Fast and reliable identification of materials is important for various applications on Earth, including geological prospecting, material and engineering sciences, and other analytical studies. For planetary surface exploration, particularly missions involving sample return, obtaining *in-situ* information on rock mineral composition is of critical importance due to time and weight constraints. Spectroscopy is an essential analytical technique for achieving this, allowing for the structural, chemical and functional properties of planetary materials to be identified. Many robotics planetary exploration missions already take advantage of data provided by different spectroscopic methods. While early robotics landers were equipped with a single analytical instrument (Blake, 1999, e.g., Viking, Venera, Mars Pathfinder) several currently operating rovers, such as the Mars Curiosity rover, have multiple spectrometers (Wiens et al., 2012). The upcoming ESA ExoMars 2020 (Rull et al., 2017; Bibring et al., 2017) and NASA Mars 2020 (Wiens et al., 2017) landers will also be equipped with analytical laboratories that are able to obtain multi-spectral data-sets from instruments such as Raman scattering (Raman), Visible-Infrared reflectance, Laser-Induced Breakdown (LIBS) and other types of spectrometers.

The ESA-PANGAEA Mineralogical Toolkit[2] aims to enhance the recognition of planetary minerals through the development of an effective analytical software suite for identifying minerals in-situ based on their mineralogical composition (that can be inferred with Raman and VNIR spectroscopy) and their chemical abundances (utilizing LIBS or other techniques). It also seeks to compile a custom multispectral data library for all known minerals present on the Moon, Mars, and other planetary bodies. Developed and tested together, this analytical data and software suite is conceived as a real-time decision support tool for future human and robotic planetary surface exploration missions(Sauro et al., 2018; Drozdovsky et al., 2019). To contribute to this goal, in this work we investigate a Machine Learning (ML) based approach for recognition of minerals based on their mineralogical composition (that can be inferred with Raman and VNIR spectroscopy) and their chemical abundances (utilizing LIBS or other techniques).

---

Various advanced ML algorithms have been proved to allow relatively fast and accurate supervised classifications. The ML classification accuracy can be progressively improved by adding new training data to the classification models without losing the recognition speed. Motivated by these benefits, in this study we investigate if a classification algorithm, "classifier", based on data from pair of spectroscopic methods (Raman, VNIR, LIBS) can achieve higher mineral identification accuracy than a classifier that uses a single spectroscopic technique. As has been demonstrated, when combining classifiers, it is particularly useful if they produce errors in an uncorrelated manner (Ali, 1995). This can be achieved by using two types of data with complementary information about the patterns to be classified (Ali, 1995; Simonyan and Zisserman, 2014a). In spectroscopy, it has been shown that different analytical techniques, such as Raman, VNIR, and LIBS we are evaluating, provide complementary information to each other. Specifically, the combination of LIBS together with hyperspectral images in VNIR/SWIR range (Haavisto et al., 2013) as well as combining Raman spectroscopy with Laser-induced fluorescence (Kauppinen et al., 2014) or LIBS (Rammelkamp et al., 2019) spectra have been favorably evaluated.

In this paper, we present a novel approach for automatic mineral classification that combines data from different spectroscopic methods and evaluates it against large datasets. We also assess different data fusion methods and ensemble techniques. We first describe machine learning methods used to identify mineral species using solely data from stand-alone analytical methods, Raman, VNIR or LIBS, and then introduce our approach: combination of pairs of the mentioned spectroscopic methods. In each section, we report mineral classification accuracy of the ML methods on spectra obtained from open access databases that are evaluated via cross-validation ("out-of-sample") technique (Geisser, 1993; Kohavi, 1995; Hastie et al., 2009).

## 2. Stand-alone analytical methods

### 2.1. Raman

Machine learning methods have found many applications in the analysis of Raman spectra, including the identification of various chemical compounds. Several different classifiers have been used to recognize the minerals, including Extremely Randomised Trees (Sevetlidis and Pavlidis, 2019, Method 1), weighted-neighbors classifier (Carey et al., 2015b, Method 2), or Convolutional Neural Networks (Liu et al., 2017, Method 3). In this study, we evaluated each of these three classification approaches and decided to introduce two additional novel methods as described below.

We assessed different classification methods based on the same dataset of Raman spectra. The previous state-of-the-art classification methods were tested on different versions of the constantly evolving open-access RRUFF Raman spectra database (Lafuente et al., 2016) and therefore used datasets with different amounts of spectra. They also used different processes to split the data into training and testing sets and used different augmentation techniques to increase the size of the training set. Liu et al. (2017) reports accuracy 88.4% and Sevetlidis and Pavlidis (2019) reports accuracy 88.8%.

To compare these state-of-the-art methods under identical conditions, a static version of RRUFF was kindly provided to us by C. J. Carey. This was the same dataset used in his work on mineral classification (Carey et al., 2015b). The dataset consists of 3950 spectra from 1214 different mineral species.

Typically, Raman spectra are preprocessed before these classifiers are applied (Carey et al., 2015a,b; Sevetlidis and Pavlidis, 2019) in order to help in the elimination of noise (unwanted signals) and to enhance discriminating features. As was done in Carey et al. (2015b), we split the dataset into a "training set", constructed by selecting three spectra per mineral species at random, and assigning the remaining spectra into the testing set. For the training set, we remove outliers from all available Raman spectra of each mineral by finding the average spectrum for each class and subsequently removing spectra that have cosine distance from the average spectrum higher than 0.5. Outlier removal is performed to ensure that the training set is not skewed by spectra affected by random instrumental artifacts or due to misclassification of the sample.

Before feeding the data into neural networks, we perform linear interpolation of these data to convert each spectrum to a vector of 1715 intensity values, sampling uniformly from 85 to 1800 $cm^{-1}$ following (Carey et al., 2015b). Finally, we normalize the intensities of each spectrum to the [0, 1] range.

*Accuracy testing of five ML methods*

The processed RRUFF dataset was used to evaluate mineral identification capabilities of the following ML classification methods:

1. Extremely Randomised Trees proposed by Sevetlidis and Pavlidis (2019)

2. Weighted-neighbors classifier by Carey et al. (2015b)

3. CNN method proposed by Liu et al. (2017)

### OUR PROPOSED METHODS

In addition to these three state-of-the-art methods, we also employed two novel methods which to our knowledge have never before been used for the mineral classification:

4. Running averages of trained variables (Averages)

5. Ensemble of different network models (Ensemble)

Method 4 improves classification accuracy of the CNN described by Liu et al. (2017) by using running averages of the trained variables instead of the values from the last training step. Namely, we used an exponential decay with decay rate of 0.999.

Method 5 is an ensemble of 6 different neural network models (all of them use running averages), with different architectures. The first two architectures were variations of the CNN described by Liu et al. (2017). The next two architectures use variation of an architecture focused on rich feature representation of the input through the use of Parallel Feature Extraction Blocks (FeatEx; Burkert et al., 2015). The last two architectures, $5th$ & $6th$, use variations of a standard convolutional network "VGG" net (Simonyan and Zisserman, 2014b). We used simpler versions of the VGG net consisting of 6 convolutional layers and 2-3 "fully connected layers"[3] which allowed us to test the ensemble over multiple independent runs.

For easier reproducibility, we report accuracies of these methods without any use of data augmentation. In Table 1 we report the mineral classification accuracies and 95% confidence interval (CI) of the accuracies over 30 independent runs. Note that, we do not report 95% confidence interval for the second method as it was not provided in the original paper. Accuracy is defined here as the percentage of mineral spectra that were correctly recognized within its class of species.

**Table 1**
Accuracies of the compared Raman classification methods

| Method | Accuracy | 95% CI |
|---|---|---|
| Method 1 - Extr. Randomised **Trees** (Sevetlidis and Pavlidis, 2019) | 80.92% | ±0.48 |
| Method 2 - Weighted-neighbors classifier (Carey et al., 2015b) | 84.80% | - |
| Method 3 - CNN (Liu et al., 2017) | 86.34% | ±0.50 |
| Our Method 4 - CNN running averages | 87.93% | ±0.48 |
| Our Method 5 - Ensemble of 6 architectures | **89.31%** | ±0.33 |

### *Testing Data Augmentation*

To further investigate the effects of data augmentation on Raman spectra, we evaluated different augmentation techniques on the latest version of the RRUFF database from September 15th 2019. From a total of 27078 spectra, we selected only 'excellent' quality data without overwhelming specimen fluorescence which were independently confirmed via the XRD analysis and have been already processed to have the baseline corrected and instrumental artifacts removed resulting in a subset containing 8950 spectra representing 1705 mineral species. The training set was created by using three spectra per mineral species at random. The remaining spectra, including spectra of a lower quality, were used for testing.

We have tested several augmentation techniques. We test the effects of shifting each spectrum left or right a few wavenumbers randomly, adding single random value to each intensity value of a single spectrum or adding random noise, proportional to the magnitude at each wavenumber. Furthermore, we evaluated augmentation proposed by Jannik Bjerrum et al. (2017), which has until now only been tested on NIR spectra. Apart from applying domain-specific transformations, we test the effects of a domain-agnostic method called "Synthetic Minority Over-sampling

---

[3] Every node in the first layer is connected to every node in the second layer

Technique" (SMOTE) (Chawla et al., 2002). We used every augmentation technique to double the size of the training samples for each class.

To evaluate the augmentation techniques comprehensively we test their effects on mineral classification accuracy with different types of classifiers. Namely, we use convolutional neural network (CNN), k-nearest neighbors algorithm (KNN), support-vector machine (SVM), and extremely randomized trees (Trees). In Table 2 we also report an average accuracy of all four classifiers. The augmentations were tested over 30 independent runs.

We did not find the effects of any examined data augmentation technique on this particular dataset to improve the classification with statistical significance. We believe this could be explained by a small intra-class variance of the dataset.

**Table 2**
Effects of different augmentation techniques

| Augmentation Technique | CNN | KNN | SVM | Trees | Avg |
|---|---|---|---|---|---|
| No augmentation | 76.90 | 68.48 | 78.31 | 67.13 | 72.70 |
| Add random value | 76.17 | 69.03 | 78.64 | 64.19 | 72.00 |
| Shift spectrum | 74.77 | 69.01 | 76.93 | 68.89 | 72.4 |
| Noise | 76.95 | 69.30 | 78.30 | 68.47 | 73.25 |
| SMOTE | 76.54 | 68.44 | 77.84 | 67.69 | 72.62 |
| Offset, slope, multiply | 76.62 | 69.36 | 78.33 | 69.36 | 73.41 |

## 2.2. VNIR

Reflectance (*aka* 'absorption') spectrometry in UV-Visual-IR wavelengths (VNIR) is another important, complementary to the Raman technique that is sensitive to chemical and physical properties and delivers a wide range of information about the analyzed sample. However, in order to extract the information, multivariate calibration of the spectral data is said to be required (Tomuta et al., 2017). With the emergence of approaches based on using supervised neural networks trained on available spectroscopic database, this task could be potentially simplified allowing to directly infer mineral and geochemical classification from unknown spectra (Tanabe et al., 1994; Parakh et al., 2016; Tanaka et al., 2019, and others).

*Preprocessing*

For the VNIR-reflectance spectra, we follow the similar pre-processing steps as it is described in the Raman section including the linear interpolation of the VNIR spectra within the wavelength range from 350 nm till 4000 nm and the normalization of the spectral reflectivity values to [0, 1] range before performing classification. Baseline (continuum) subtraction was not applied to the VNIR spectra.

*Accuracy Testing*

Similarly to the Raman spectroscopy, we evaluated the accuracy of our method of classification of VNIR spectra based on the ensemble of 6 convolution networks. To create training and testing set, we combined spectra from open access databases RELAB issued on December 31st 2019 (NASA Reflectance Experiment Laboratory), USGS version 7 (Kokaly et al., 2017), and ECOSTRESS version 1.0 (Meerdink et al., 2019)(Baldridge et al., 2009). The final dataset comprised of 6231 spectra, representing 366 different mineral species (less than a quarter of all minerals with Raman spectra!). We split the dataset into training and testing sets using a leave-one-out scheme by selecting random spectra per mineral species for the testing set and assigning the remaining spectra into the training set. We performed 30 independent runs and report average mineral classification accuracy 69.71% with a $\pm 0.71$ 95% confidence interval. Again, the CNN ensemble provided the highest accuracy.

## 2.3. LIBS

The LIBS technique probes the atomic chemical composition of the material rather than its molecular structure as the VNIR and Raman spectroscopy. Nonetheless, the chemical composition is potentially valuable complementary information allowing to distinguish minerals and rocks when used in combination with other analytical methods. We have implemented two different LIBS algorithms for elemental estimation, both relying on the standard emission lines library of 81 chemical elements from the NIST LIBS Database (Kramida et al., 2017). The first algorithm, inspired by text retrieval, finds spectral peaks of the queried sample and the theoretical spectral peaks of each atomic element (obtained from NIST) and afterwards represents the normalized peak intensities as weighted vectors (Amato et al., 2010). Then, the algorithm estimates chemical composition by computing cosine similarity between the queried weighted vector and the weighted vector each atomic element. We then make mineral classification predictions by comparing the calculated elemental composition of each mineral from its empirical formula (taken from webmineral.com) and the predicted mineral elemental composition. The second algorithm uses CNN that was trained on a synthetic dataset created from theoretical LIBS spectra of random elemental compositions. The algorithms were compared by predicting the elemental composition of minerals containing elements that occur naturally on Earth (albeit from larger, first 94, elements of the periodic table).

The CNN outperformed the cosine similarity algorithm by achieving a significantly lower mean absolute error (MAE) of the predicted elemental composition (0.156 vs 0.174) and higher mineral classification accuracy (8.98% vs 6.44%).

## 3. Pair-combined analytical methods

### 3.1. Methodology

The methods used to combine Raman or VNIR together with LIBS differ from the methods used to combine VNIR and Raman data. When combining Raman and VNIR, we either trained two separate classifiers that predicted mineral species and then combined these predictions or used a single two-stream convolutional neural network Simonyan and Zisserman (2014a). On the other hand, we used LIBS data to estimate elemental composition and subsequently fused this information with the VNIR/Raman prediction to classify the mineral species. Flow diagram of our fusion approach of the Raman/VNIR and LIBS spectra for recognition of minerals is presented on the Fig. 1. In order to evaluate the robustness of the combination of any two spectroscopic methods and simulate more natural conditions, in the following sections we do not exclude lower-quality spectra. To save computational time, whenever we used CNN to test any combination of data obtained by different spectroscopic methods, we used simple architecture with 4 convolutional layers and 2 fully connected layers, without any data augmentation and without using the exponential weighted average of the trained variables. We used dropout regularization to prevent overfitting of the CNN.

### 3.2. Raman + VNIR

When recognizing both Raman spectra and VNIR spectra, we used CNN as a classifier. We experimented with different ensemble techniques and took two different general approaches.

The first approach consisted of training two different classifiers, one for Raman spectra and second for VNIR spectra, and then combining predictions (softmax scores) of each classifier by late fusion (Simonyan and Zisserman, 2014a). We experimented with three fusion methods: (i) averaging the predictions (Ave-p), (ii) multiplying the predictions (Mul-p), or (iii) having a support vector machine (SVM) to learn the relationship between the predictions of both classifiers and the correct label. Note, that the $i$-th component of the fused vector (by multiplication) is the result of multiplication between the $i$-th component of the Raman/VNIR prediction and $i$-th component of the LIBS prediction.

The second approach was using a single CNN with two separate recognition streams (Raman and VNIR), that fuses the streams at the last convolutional layer (Feichtenhofer et al., 2016).

Because we do not have Raman and VNIR spectra from the same mineral sample (instance of a single mineral) we tested these approaches on an artificially created pair-combined dataset. The dataset was compiled from the same open access databases described in the previous 2.1 and 2.2 sections: Raman spectra obtained from RRUFF database and VNIR spectra from RELAB, USGS, and ECOSTRESS databases. To be specific, we only used subsets consisting of minerals found in both Raman and VNIR databases, which consisted of spectra from 259 different mineral species. We used a leave-one-out cross-validation method to split a dataset into training and test sets by randomly selecting a single spectrum per mineral type for testing and using the rest for training. We paired each Raman and VNIR spectra from the same mineral species as synthetic data points (features). In Table 3 we report mineral classification accuracies of the compared methods averaged over 30 independent training runs. To be certain that indeed combining different types of data provides the best accuracy, we also tried creating synthetic mineral samples by making pairs of two different Raman spectra from the same mineral species. On this dataset, we have a achieved mineral classification accuracy of 88.95%. Therefore, we can see that the most accurate method takes advantage of the information present in both Raman and VNIR data instead of using one type of data.

**Figure 1:** The simplified procedure of our method for recognizing minerals from combined Raman/VNIR and LIBS spectra.

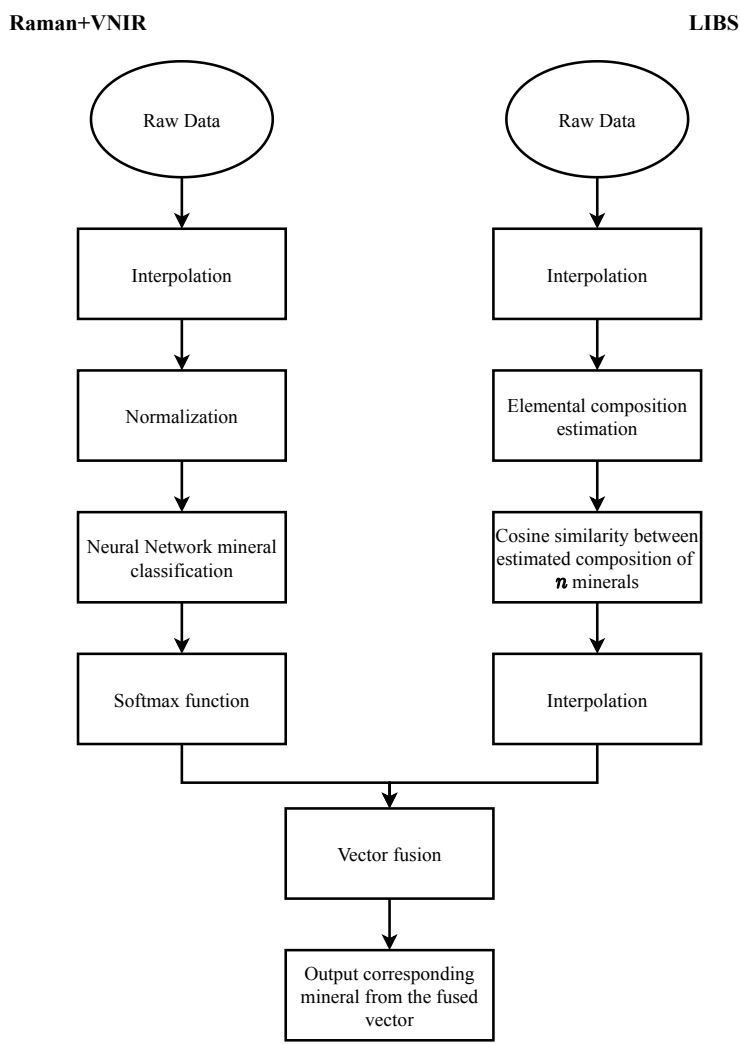

**Table 3**

Classification accuracies from the individual Raman or VNIR spectroscopy versus prediction scores of various combined Raman+VNIR spectroscopic classifications. In bold the highest accuracy obtained through Averaging the predictions (Ave-p).

| Method | Individual | | Combined VNIR+Raman | | | |
|---|---|---|---|---|---|---|
| | VNIR | Raman | Fusion | Ave-p | Mul-p | SVM |
| CNN+CNN | 76.71% | 85.38% | 85.15% | **92.76%** | 92.57% | 91.35% |

### 3.3. Raman + LIBS

To evaluate the combination of Raman and LIBS we used a subset of minerals that has Raman spectra from RRUFF and which overall chemical composition could be calculated with the NIST LIBS Database. The resulting dataset had spectra from 1165 different mineral species. We compared mineral classification accuracy of two different combinations: (i) the Raman CNN classifiers with the LIBS CNN classifier, and (ii) the Raman CNN classifier with the LIBS cosine similarity algorithm. To combine the Raman classifier with the LIBS classifier we use late fusion of predictions. Prediction of the CNN is a softmax score and in the case of the LIBS cosine similarity algorithm the prediction is a $L_1$ normalized vector $\vec{x} = x_1, x_2..x_n$, where $n$ is the number of minerals and $x_i$ is the cosine similarity between the elemental composition of $i$-th mineral and the predicted elemental composition. We experimented with three different fusion methods: (i) averaging the predictions, (ii) multiplying the predictions, and (iii) squaring the predictions (Sq-p) of the LIBS classifier before multiplying it with the predictions of the Raman classifier. In Table 4 we report the mineral classification accuracies of the compared methods averaged over 30 independent training runs.

**Table 4**

Classification accuracies from the individual Raman or LIBS spectroscopy versus prediction scores of various combined Raman+LIBS spectroscopic classifications. In bold the highest accuracy obtained through Averaging the predictions (Sq-p) of the CNN+cosine classifications fusion method.

| Method | Individual | | Combined LIBS+Raman | | |
|---|---|---|---|---|---|
| | LIBS | Raman | Ave-p | Mul-p | Sq-p |
| CNN+cosine | 6.44% | 79.04% | 80.40% | 81.92% | **83.21%** |
| CNN+CNN | 8.98% | 78.88% | 78.04% | 77.69% | 76.82% |

**Figure 2:** Violin plot: X-axis shows distributions of predictions by two different algorithms; Y-axis is the cosine similarity between the correct elemental composition and the predicted elemental composition

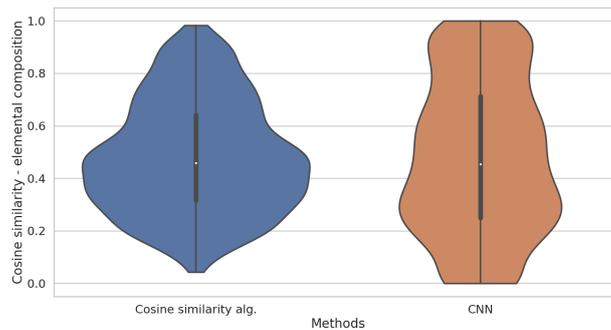

The violin plot (Hintze and Nelson, 1998), see Fig. 2, shows a full distribution of the cosine similarities between the elemental composition of queried mineral spectra and the predicted elemental composition. It can be observed that the cosine similarity algorithm had a low number of completely incorrect predictions, i.e., with low or no similarity between the elemental composition of queried mineral and the ground truth elemental composition. This characteristic property allowed us to improve the mineral classification accuracy of the Raman+LIBS combined classifier by initially squaring the prediction value of the cosine similarity algorithm before multiplying it with the Raman classifier prediction.

### 3.4. VNIR + LIBS

To evaluate the combination of VNIR and LIBS spectra for mineral classification we repeated the same procedures and methods used for Raman and LIBS combination. To train the VNIR classifier, we again combined all the available spectra from RELAB, USGS and ECOSTRESS database. We then selected a subset of the dataset composed of 279 different mineral species which are present in both VNIR and LIBS dataset. In Table 5 we report accuracies of the compared methods averaged over 30 independent training runs.

## 4. Discussion and Conclusion

### *Complementary pair-method combined spectroscopy for the mineral classification*

In this paper, we have presented a novel approach for mineral classification that combines data obtained by three different spectroscopic methods. Each of the combinations, VNIR+Raman, Raman + LIBS, or VNIR + LIBS outperformed the use of only one type of spectroscopic data. A clear advantage of different complementary spectroscopic techniques is the possibility to better distinguish variations in chemical compositions versus molecular structure.

**Table 5**
Classification accuracies from the individual VNIR or LIBS spectroscopy versus prediction scores of various combined VNIR+LIBS spectroscopic classifications. In bold the highest achieved accuracy was again through Squaring the predictions (Sq-p) of the CNN+cosine classifications fusion method.

| Method | Individual | | Combined VNIR+LIBS | | |
|---|---|---|---|---|---|
| | LIBS | VNIR | Average | Multiply | Square |
| CNN+cosine | 15.75% | 73.01% | 77.49% | 77.24% | **79.04%** |
| CNN+CNN | 9.31% | 78.53% | 78.14% | 79.34% | 78.14% |

The method which provides the highest accuracy when combining Raman+LIBS or VNIR+LIBS differs from the method used to combine VNIR and Raman data. The main difference is in the estimation of elemental composition from the LIBS spectra and subsequently, a different method that fuses this information with the prediction of Raman or VNIR classifier (Fig. 1). Although the combination of Raman and VNIR achieved the highest mineral classification accuracy, this result could be affected by the differences in the available pair-combined spectral datasets.

Nonetheless, the improvement in detection accuracy by combining the Raman scattering and VNIR absorption spectra is rather expected, as these two types of vibrational spectra are known to be complementary to each other by being excited by different and in some cases mutually exclusive vibrational transitions in molecules (Sharma, 1981; Hashimoto et al., 2019).

The complementary nature of the bespoken spectroscopic techniques can be seen for example in the following two mineral groups.

***An example of olivine solid solution series***

As can be seen in Figure 3, olivines exhibit diagnostic absorption features across visible to near-infrared (VNIR) wavelengths due to electronic transitions of $Fe^{2+}$ or Mg in their crystal structure (e.g., Isaacson et al., 2014). At the same time, the Raman spectra of the olivine-group minerals have a characteristic set of two intense lines of the Si-O asymmetric stretching band and Si-O symmetric stretching band (e.g., Mouri and Enami, 2008; Dyar et al., 2009; Breitenfeld et al., 2018). Moreover, for both of the vibrational spectroscopic methods the subtle changes in chemical composition could lead to recognizable modifications of their vibrational spectroscopic features, as it can be tracked already from the simple Principle Component Analysis, shown on the bottom plots (c,d) of Figure 3.

***Deep Convolution Neural Network for improved mineral recognition from single-method spectroscopy***

In addition, we have presented a Deep Convolutional Neural Network solution for the Raman spectrum classification that significantly outperformed previous state-of-the-art methods based on tests performed on the same Raman spectral archive.

In our follow up work Drozdovsky et al. (2020a in preparation) we are planning to use our algorithms to evaluate the best combination of analytical methods for identification of subset of rocks and minerals expected to be present on the surface of Moon, Mars and minor Solar System bodies. Additionally, we are planning to further validate our pair-methods combined classifications algorithms using the same sample of rocks and minerals which were simultaneously measured with all three analytical methods as a part of the PANGAEA-X field test campaigns (Drozdovsky et al., 2019).

## 5. Acknowledgment

P.J would like to thank the CAVES & PANGAEA Team at the ESA/EAC for providing internship opportunity that resulted in this work. This research utilizes data from the following open-access repositories: RRUFF, RELAB, ECOSTRESS, USGS, NIST-LIBS and Webmineral. We would like to thank Dr. C.J. Carey for sharing with us his version of the RRUFF Raman spectra, and Dr. N.J. McMillan for sharing her LIBS spectral archive.

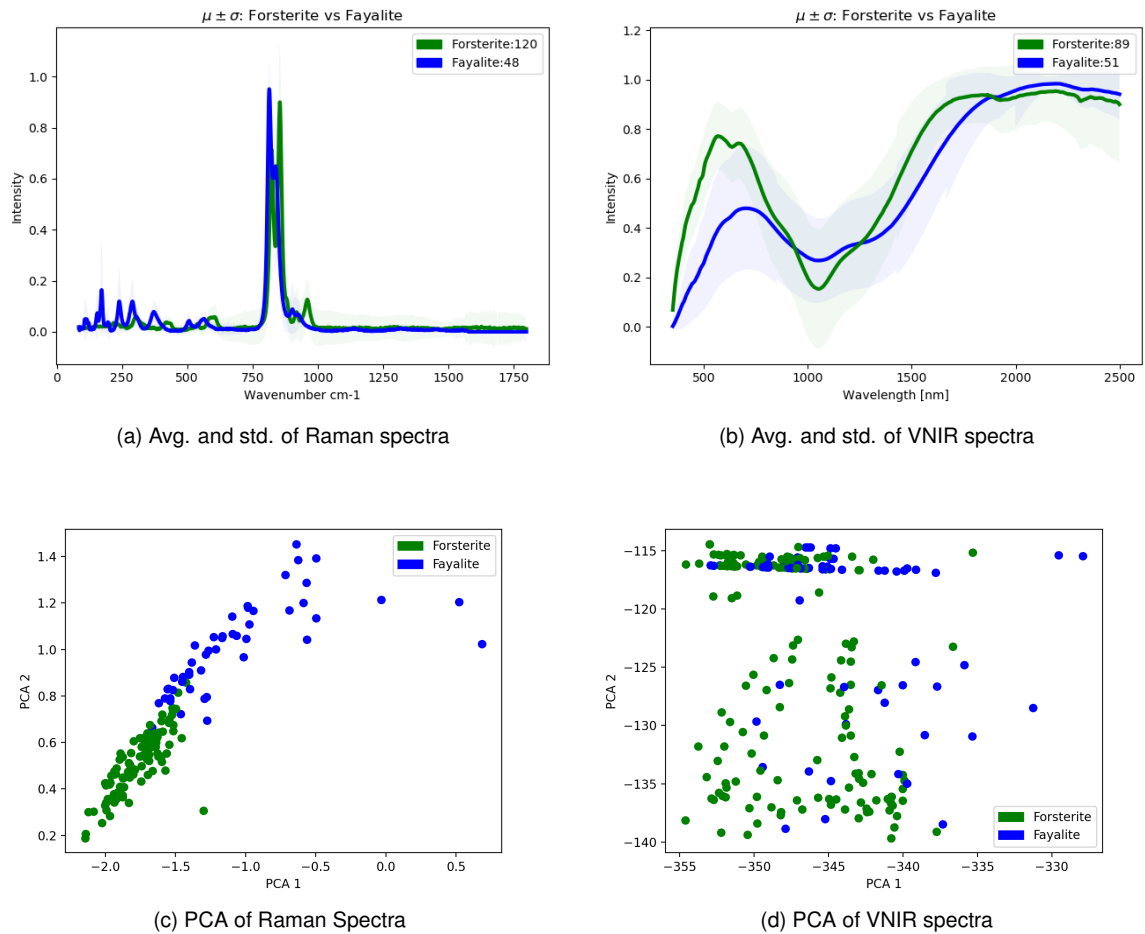

(a) Avg. and std. of Raman spectra

(b) Avg. and std. of VNIR spectra

(c) PCA of Raman Spectra

(d) PCA of VNIR spectra

**Figure 3:** Average and standard deviation of all available Raman & VNIR spectra for two end-members of olivine solution series, Forsterite (shown in green) and Fayalite (shown in blue) (a,b). (Bottom row) Scatter plot of two principal components in Raman and VNIR spectra of Forsterite (green filled circles) and Fayalite (blue filled circles) (c,d)

## Authorship contribution statement

**Pavel Jahoda:** Machine learning, Methodology, Data processing, Writing. **Igor Drozdovskiy:** Conceptualization of this study, Writing, Data curation. **Francesco Sauro:** Supervision, Review. **Leonardo Turchi:** Review. **Samuel Payler:** Review. **Loredana Bessone:** Supervision.